\documentclass[letterpaper, 10 pt, conference]{ieeeconf}

\IEEEoverridecommandlockouts                              

\IEEEoverridecommandlockouts
\usepackage{amsmath,amssymb,amsfonts}
\usepackage{graphicx}
\usepackage{textcomp}
\usepackage{xcolor}
\usepackage{float}
\usepackage{booktabs}
\usepackage{makecell}
\usepackage{array}
\usepackage{multirow}
\usepackage{tikz}

\usepackage{cite}

\makeatletter
\let\NAT@parse\undefined
\makeatother
\usepackage[
    final,
    allcolors=blue,
    colorlinks=true
]{hyperref}

\usepackage{amsmath} % assumes amsmath package installed
\usepackage{amssymb}  % assumes amsmath package installed

\def\diag{\mathrm{diag}}

\newcommand{\calV}{\mathcal{V}}

\newcommand{\R}{\mathbb{R}}

\usetikzlibrary{shapes, arrows, positioning, calc, fit, arrows.meta, decorations.pathmorphing}
\tikzstyle{every picture}+=[remember picture]
\everymath{\displaystyle}
\tikzset{
	>=stealth',
	pil/.style={red!50,very thick,dotted}
}

\usetikzlibrary{positioning, fit, calc, backgrounds, shadows, arrows.meta}

\definecolor{execdark}{HTML}{1F4E79}   % Deep Blue
\definecolor{execlight}{HTML}{DEEBF7}  % Light Blue
\definecolor{accentred}{HTML}{C00000}  % Deep Red
\definecolor{accentredlight}{HTML}{F8CBAD} % Light Red
\definecolor{accentgreen}{HTML}{A9D18E} % Soft Green
\definecolor{accentviolet}{HTML}{7030A0}      % Deep Violet

\definecolor{execdark}{HTML}{1F4E79}   % Deep Blue
\definecolor{execlight}{HTML}{DEEBF7}  % Light Blue
\definecolor{accentred}{HTML}{C00000}  % Deep Red
\definecolor{accentredlight}{HTML}{F8CBAD} % Light Red
\definecolor{accentgreen}{HTML}{548235} % Deep Green
\definecolor{accentgreenlight}{HTML}{E2EFDA} % Light Green
\definecolor{accentvioletlight}{HTML}{E4DFEC}  % Light Violet

\definecolor{violet}{RGB}{88, 66, 155}
\definecolor{darkBlue}{RGB}{0, 51, 102}       % deep blue
\definecolor{darkGreen}{RGB}{0, 102, 51}      % forest green
\definecolor{darkRed}{RGB}{153, 0, 0}         % crimson/burgundy
\definecolor{darkCyan}{RGB}{0, 102, 102}      % teal/cyan
\definecolor{darkMagenta}{RGB}{102, 0, 102}   % deep magenta

\definecolor{orangeRed}{RGB}{237, 19, 90}

\begin{document}

\title{\LARGE \bf Robust Nonprehensile Object Transport with Quadruped Robots}
\author{Ainoor Teimoorzadeh$^{*}$, Riccardo Pretto$^{*}$, Mario Selvaggio, Gokhan Alcan, Sami Haddadin
\thanks{$^{*}$These authors contributed equally to this work.}
\thanks{A. Teimoorzadeh is with Munich Institute of Robotics \& Machine Intelligence, Technical University of Munich (TUM), Munich, Germany, E-mail:  {\tt\small ainoor.teimoorzadeh@tum.de}}
\thanks{R. Pretto and G. Alcan are with the Tampere University, Finland. {\tt\small\{riccardo.pretto, gokhan.alcan\}@tuni.fi}}
\thanks{M. Selvaggio is with PRISMA Lab, Department of Electrical Engineering and Information Technology, University of Naples Federico II, Via Claudio 21, 80125, Naples, Italy. E-mail: {\tt\small mario.selvaggio@unina.it}}
\thanks{S. Haddadin is with Mohamed Bin Zayed University of Artificial Intelligence, Masdar City, Abu Dhabi, UAE. E-mail: {\tt\small sami.haddadin@mbzuai.ac.ae}}
\thanks{We gratefully acknowledge the support of the Lighthouse Initiative Geriatronics, funded by LongLeif GaPa gGmbH (Project Y, grant no. 5140953), the Italian Ministry of University and Research (MUR) under the Fondo Italiano per la Scienza (FIS)—FIS 3 (ROMOTIVE Project), as well as NVIDIA Corporation for their support to Tampere University for this work through the Academic Grant Program.}
}

\maketitle

\begin{abstract}
In this paper, we present a robust nonprehensile object transportation framework for quadruped robots. An uncertainty-aware trajectory optimization method generates object motions with minimal closed-loop sensitivity to uncertain parameters.
The resulting reference trajectory is tracked using a coupled convex model predictive controller that jointly predicts the CoM dynamics of the quadruped and the payload followed by a whole-body QP that enforces ground reaction constraints. 
The approach is evaluated through extensive simulations and real-world experiments under variations in the object’s inertial parameters. 
Its performance is compared with fixed-orientation and straight-line trajectories as baseline. 
The results show that the optimized object motion reduces the sliding  by
approximately $50\%$ compared with the fixed-orientation baseline and
$30\%$ compared with the straight-line baseline, while also achieving
lower robot CoM tracking errors.
\end{abstract}

\section{Introduction} \label{sec:introduction}
Manipulation is a core capability that robots need to acquire to become helpful human assistants across various domains~\cite{trobinger2021introducing, srinivasa2012herb, Kempdesign}. 
Nevertheless, most existing manipulation approaches still rely on form- or force-closure~\cite{bicchi1995closure}, often requiring sophisticated and fragile grasping operations. 
Dynamic nonprehensile manipulation~\cite{lynch1999dynamic} has broadened the range of robotic manipulation applications by offering advantages such as controlling more degrees of freedom than the manipulator itself possesses~\cite{ruggiero_nonprehensile}, or widening its workspace through human-like manipulation actions such as tossing and catching~\cite{satici2016coordinate}.
Dynamic nonprehensile object transportation, also known as friction-based manipulation or dynamic grasp, keeps a non-grasped object in contact with a tray-like end-effector throughout the robot motion~\cite{Flores2013}. 
Some existing approaches, which realize this manipulation primitive by enforcing non-sliding conditions, include shared control with haptic feedback~\cite{selvaggio2021shared}, Quadratic Programming (QP) based trajectory optimization methods~\cite{subburaman2023non}, and Model Predictive Control (MPC) frameworks~\cite{Heins_2023, selvaggio2023non}.

However, most of the existing approaches are designed for dexterous robotic arms, which are well-established platforms for such tasks~\cite{haddadin2024franka}. 
Although robotic arms can execute fast and precise motions while keeping the object in nonsliding conditions~\cite{pham2016admissiblevelocitypropagation, 10403905}, their workspace is limited by their reach.
Mobile manipulators overcome this limitation by combining the extended workspace of a mobile base with the manipulation capabilities of an arm~\cite{Heins_2023, Heins_2025}. 
Legged robots further extend this mobility to complex environments, such as stairs and uneven terrain, that may be inaccessible to wheeled platforms. 
Yet, keeping non-grasped objects fixed to an actuated legged platform comes with additional challenges: dynamic phenomena, such as high-frequency chassis vibrations, intermittent impacts from the trotting gait, cause unpredictable friction variations at the object's interface.
Moreover, the presence of an object alters the robotic system's center of mass and causes unexpected inertial shifts, making precise trajectory tracking a major control challenge motivating the adoption of robust control techniques~\cite{teimoorzadeh2022robust}.
\begin{figure}[t]
	\centering
	\includegraphics[width=0.8\linewidth]{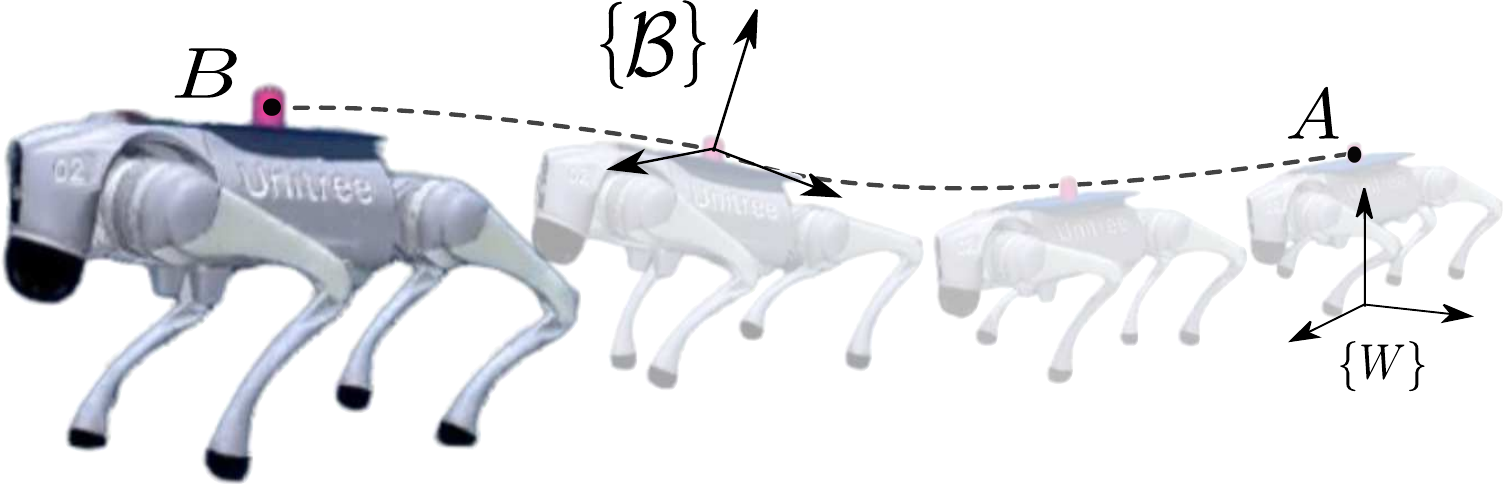}
	\caption{Picture of the problem addressed in this paper: A quadruped carries an unrestrained object on its back mounted flat platform from the initial point $A$ to the final point $B$ over an optimized sensitivity-aware trajectory planned for the reference frame $\{\mathcal{B}\}$ attached to the object.
	\label{fig:concept}
    \vspace{-0.1cm}
	}
\end{figure}

In contrast to the well-established robust control methods, such as adaptive control~\cite{aastrom1995adaptive}, passivity based control~\cite{ortega2004interconnection}, $H_{\infty}$ or $\mu$-synthesis~\cite{zhou1998essentials}, 
closed-loop sensitivity approach allows a dynamical system to fulfill the control objective by designing control-aware trajectories whose execution is minimally sensitive to physical parameter uncertainties~\cite{giordano2018trajectory}. 
Despite its promising results in domains such as aerial manipulation~\cite{srour2024experimental, belvedere2025sensitivity}, the approach remains largely unexplored in nonprehensile manipulation settings, with only limited work reported to date~\cite{teimoorzadeh2025robust}.

To address this gap, we present an integrated planning and control
framework for robust nonprehensile object transportation using a
quadrupedal robot.
We consider a cylindrical object which is transported on a back-mounted plate of a quadruped robot. 
The object is constrained only by friction on one of its circular faces, making it free to slide relative to the support surface. 
While this configuration bypasses the mechanical and control complexities of integrating a dedicated robotic arm~\cite{morlando2022nonprehensile}, it introduces significant transport challenges, as the sliding object can only be indirectly prevented through the regulation of the robot motion itself.

The proposed framework, therefore employs closed-loop sensitivity-based trajectory optimization to generate object-centric
motions which are tracked using a coupled robot-object MPC and a whole-body controller.

In summary, the main contributions of this work are:
\begin{enumerate}
    \item Formulation of a closed-loop sensitivity optimization to generate object-centric motions for an unsecured payload carried directly on a quadruped.
        
    \item  A tailored coupled quadruped-object convex MPC
    with whole-body control for nonprehensile transportation on a back-mounted platform, extending previous arm-based quadruped approaches.

    \item Validation of the proposed approach through simulation in MuJoCo environment and hardware experiments, showing reduced object sliding compared with baseline trajectories.

\end{enumerate}

\subsection{Related works}

Nonprehensile object transportation has been extensively studied for fixed-base and mobile manipulators. Early approaches formulate the waiter’s problem as a trajectory-planning problem and generate motions that satisfy contact-stability constraints or optimize execution time~\cite{Flores2013, CSORVASInearoptimal, pham2016admissiblevelocitypropagation, GATTRINGER2023105286, 10403905}.
These methods, however, generally assume accurate knowledge of the object and contact properties and evaluate feasibility using nominal models. 
Consequently, trajectories that are feasible for the assumed parameters may become fragile when the actual mass distribution, inertia, or friction differs from its nominal value.
Feedback-based methods partially alleviate this limitation by adapting the robot motion during execution. 
The MPC formulation in~\cite{selvaggio2023non} explicitly accounts for object dynamics, frictional contact, and robot constraints, but relies on a known object model.

In~\cite{jawale2024dynamicnonprehensileobjecttransport}, MPC is augmented with value functions learned from demonstrations, enabling empirical generalization to previously unseen objects, learning robustness from data.
More recently, the work in~\cite{Heins_2025} introduced robust sticking constraints for uncertain CoM and inertial parameters. That approach plans motions that remain feasible 
over a prescribed uncertainty primarily addressing robustness at the planning level. 

Sensitivity-aware planning and control constitutes an alternative approach. Minimum-sensitivity trajectory generation was introduced in~\cite{giordano2018trajectory} and experimentally validated on aerial robots in~\cite{srour2024experimental}, while~\cite{belvedere2025sensitivity} incorporated state sensitivity into MPC for robotic systems with parametric uncertainty. These works demonstrate that nominal performance alone is insufficient to characterize robustness: trajectories with comparable nominal behavior can produce substantially different deviations under parameter errors. The closed-loop sensitivity framework was applied to nonprehensile manipulation in~\cite{teimoorzadeh2025robust} to reduce the effect of inertial-parameter errors on the object motion. However, its validation is likewise limited to manipulation platforms whose supporting surface can be commanded independently of locomotion. Nevertheless, the use of this principle to jointly plan and control the motion of an unsecured payload carried by a legged robot remains largely unexplored.

To transport an object using a quadruped robot, the work in~\cite{morlando2022nonprehensile} equipped it with an arm and combined a momentum-based observer with a QP-based whole-body controller to compensate for external disturbances.
Although this architecture compensates for external disturbances, it requires an additional manipulator, increasing the system’s mass, mechanical complexity, and energy consumption.
LocoTouch~\cite{lin2025locotouchlearningdynamicquadrupedal}, instead, uses the robot’s back directly as a carrying surface and employs distributed tactile sensing and a learned locomotion policy to transport unsecured cylindrical objects. Its focus is on tactile-reactive locomotion—particularly for cylinders whose rolling dynamics can be directly detected through the back-mounted sensor—rather than model-based robustness to uncertain object and contact parameters.

Using the quadruped’s dorsum as a carrying surface introduces a coupling that is absent from conventional waiter-type manipulation. The platform motion is not generated by a rigidly controlled manipulator alone but results from the interaction between locomotion, intermittent ground contacts, body stabilization, and payload dynamics~\cite{morlando2022nonprehensile}. Gait-induced body oscillations and foot-impact disturbances can excite sliding or tipping of the unsecured object, while errors in its mass distribution, inertia, or contact properties alter both the predicted payload motion and its interaction with the robot. Thus, robustness cannot be obtained solely by enforcing nominal contact constraints or by rejecting disturbances at the whole-body-control level~\cite{Heins_2025}.

In contrast to the existing literature, this work combines sensitivity-aware trajectory planning with MPC-based execution for an unsecured object placed directly on a quadruped’s dorsal carrying surface. The planned motion is selected not only for nominal task feasibility but also to reduce the propagation of object-parameter uncertainty into the closed-loop state. The MPC subsequently tracks this sensitivity-aware reference while accounting online for payload dynamics and contact constraints. This combination addresses a previously unexplored problem: exploiting the quadruped body as a lightweight, manipulator-free carrying platform while explicitly shaping the robustness of nonprehensile transport against parametric uncertainty and gait-induced disturbances.

We consider the case of a cylindrical object resting with its circular base in contact with flat surface rigidly mounted on the back of a quadruped robot. 
Our aim is to compute rest-to-rest trajectories that, when executed via closed-loop control, transport the object between two prescribed points (as shown in~Fig.~\ref{fig:concept}) and exhibit minimal sensitivity to variations in the object's dynamic parameters, including friction with the robot surface.

The rest of this paper is structured as follows. 
Preliminaries and system modeling are explained in Sec.~\ref{sec:System Modeling} in a nutshell. 
Robust trajectory optimization and controller are presented in Sec.~\ref{sec:method}.
Finally, the simulation results of the closed-loop manipulation task and preliminary experimental results are shown in Sec.~\ref{sec:simulation}
Conclusions are presented in~\ref{sec:conclusions}.

\section{Preliminaries \& Object Modeling} \label{sec:System Modeling}
This section provides the minimal background on closed-loop sensitivity in Sec.~\ref{sec:cls} and on the adopted object and friction model in Sec.~\ref{sec:Object Modeling} to support the explanation of our methodology.

\subsection{Closed-Loop Sensitivity}\label{sec:cls}

To quantify robustness against parametric uncertainty, consider a generic system dynamics as
\begin{equation}
\label{eq:model}
    \dot{x}=f(x,u,p),
\end{equation}
where $x\in\R^{n_x}$, $u\in\R^{n_u}$, and $p\in\R^{n_p}$ denote
the state, input, and parameters, respectively. 
We consider that parameters $p$ may be uncertain. 
In our nonprehensile transportation case, $p$ include the object inertial parameters and friction coefficient.
Given a desired reference trajectory $y_d(t)$, a generic trajectory tracking controller described by
\begin{equation}
\label{eq:input}
    \dot{\xi}=g(\xi,x,y_d,p_n,t),\qquad
    u=h(\xi,x,y_d,p_n,t),
\end{equation}
where $\xi$ is the controller state and $p_n$ denotes the nominal parameter vector (the controller has no knowledge of the true parameters).

The closed-loop state and input sensitivities are defined as
\begin{equation}
    \Pi=\left.\frac{\partial x}{\partial p}\right|_{p=p_n}\in \R^{n_x \times n_p},
    \qquad
    \Theta=\left.\frac{\partial u}{\partial p}\right|_{p=p_n} \in \R^{n_u \times n_p},
\end{equation}
and denote variation in the system state and control, respectively, due to variation in the system parameters.
They are obtained by integrating
\begin{equation*}
    \begin{cases}
        \displaystyle\dot\Pi(t) = \frac{\partial f}{\partial x}\Pi + \frac{\partial f}{\partial u}\Theta + \frac{\partial f}{\partial p}, \quad \Pi(t_0) = 0\\[0.3cm]
        \displaystyle\dot\Pi_\xi(t) = \frac{\partial g}{\partial x}\Pi + \frac{\partial g}{\partial \xi}\Pi_\xi, \quad \Pi_\xi(t_0) = 0\\[0.3cm]
        \displaystyle \Theta(t) = \frac{\partial h}{\partial x}\Pi + \frac{\partial h}{\partial \xi}\Pi_\xi\\
    \end{cases}.
\end{equation*}
These matrices quantify how robust a system evolution is with respect to parameters' variation and can be used to optimize reference trajectories~\cite{giordano2018trajectory} (or estimate uncertainty tubes~\cite{afifi2024safe}) in order for the perturbed system evolution to exhibit minimal variation (or enforce physical bounds) with respect to uncertain parameters. 

Assume that each uncertain parameter $p_j$ varies around its nominal value $p_{c,j}$ according to
\begin{equation}
\label{eq:p_i}
    p_j\in
    [p_{c,j}-\delta p_j,\,
     p_{c,j}+\delta p_j].
\end{equation}
Defining $\Delta p=p-p_c$, the parameter uncertainty within these bounds is represented by the ellipsoid
\begin{equation}
\label{eq:ell_p}
    \Delta p^\top W^{-1}\Delta p\leq1,
    \qquad
    W=\diag(\delta p_1^2,\ldots,\delta p_{n_p}^2).
\end{equation}
Using the first-order approximations $\Delta x\approx\Pi\Delta p$ and $\Delta u\approx\Theta\Delta p$, the corresponding state and input
ellipsoids have the shape matrices
\begin{equation}
\label{eq:ell_q}
    Q_x=\Pi W\Pi^\top,
    \qquad
    Q_u=\Theta W\Theta^\top.
\end{equation}
The radius of the input ellipsoid along a direction $n_i$ is therefore
\begin{equation}
\label{eq:radius}
    r_i=\sqrt{n_i^\top\Theta W\Theta^\top n_i}.
\end{equation}
By selecting $n_i$ as the unit vector associated with the $i$-th input, the corresponding first-order uncertainty envelope is
\begin{equation}
\label{eq:tube}
    u_{c,i}-r_i\leq u_i\leq u_{c,i}+r_i,
\end{equation}
where $u_{c,i}$ denotes the nominal input. The same approach can be applied to the state or to any quantity for which a sensitivity matrix
is available.

\subsection{Object Modeling}\label{sec:Object Modeling} 
\begin{figure}[t]
	\centering
	\includegraphics[width=0.4\linewidth]{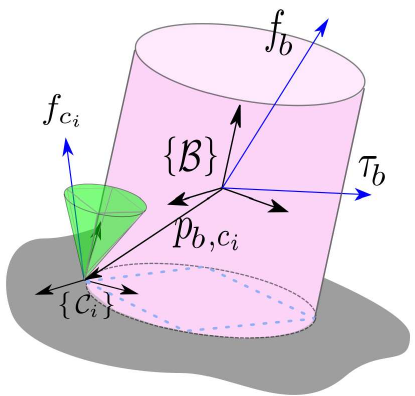}
	\caption{Schematic of the tray-based nonprehensile transportation model, in which a cylindrical object rests on a platform mounted on the back of a quadrupedal robot.
    The symbols are defined in Sec.~\ref{sec:Object Modeling}.
	\label{fig:system-description}
	}
\end{figure}
The transported object as shown in Fig.~\ref{fig:system-description} is modeled as a rigid body. Let ${\mathcal{B}}$ denote a body-fixed frame located at the object's center of mass. The object dynamics expressed in body frame are given by
\begin{equation}
M_b\dot{\mathcal{V}}_b
+ C_b(\mathcal{V}_b)\mathcal{V}_b
+ N_b(R_{s,b})
= \mathcal{F}_b,
\label{eq:object_dynamics}
\end{equation}
where $M_b\in\mathbb{R}^{6\times6}$ is the constant positive-definite mass matrix constructed from the object mass $m>0$ and the symmetric inertia matrix $I\in\mathbb{R}^{3\times3}$.
The matrix $C_b(\mathcal{V}_b)\in\mathbb{R}^{6\times6}$ accounts for
Coriolis and centrifugal effects, while
$N_b(R_{s,b})\in\mathbb{R}^{6}$ represents the gravitational wrench.
The object twist and wrench are defined as
\begin{equation}
\label{vel_torque_payload}
    \mathcal{V}_b=
    \begin{bmatrix}v_b^\top&\omega_b^\top\end{bmatrix}^{\top},
    \qquad
    \mathcal{F}_b=
    \begin{bmatrix}f_b^\top&\tau_b^\top\end{bmatrix}^{\top},
\end{equation}
where $v_b$ and $\omega_b$ are the linear and angular velocities, while
$f_b$ and $\tau_b$ are the applied force and torque.

The object-platform interaction is described by fixing $n_c=4$ point contacts with friction~\cite{prattichizzo2016grasping} along the perimeter of the circular object surface. 
This simplified contact model allows us to adopt the same consolidated description of the interaction as in~\cite{selvaggio2023non, Heins_2025} and  
constitute a conservative approximation of the true contact evolution.
The known pose of contact
frame $\{C_i\}$ relative to $\{\mathcal{B}\}$ is denoted by $q_{b,c_i}=(p_{b,c_i},R_{b,c_i})\in\mathrm{SE}(3)$. Since each contact transmits a force $f_{c_i}\in\mathbb{R}^3$ but no moment, the resulting body wrench is
\begin{equation}
\label{eq:grasp_mapping}
    \mathcal{F}_b=GF_c,
    \qquad
    G=
    \begin{bmatrix}
        \mathrm{Ad}^{\top}_{q^{-1}_{b,c_1}}B_c &
        \cdots &
        \mathrm{Ad}^{\top}_{q^{-1}_{b,c_{n_c}}}B_c
    \end{bmatrix},
\end{equation}
where
\begin{equation}
    F_c=
    \begin{bmatrix}
        f_{c_1}^{\top}&\cdots&f_{c_{n_c}}^{\top}
    \end{bmatrix}^{\top}
    \in\mathbb{R}^{3n_c}.
\end{equation}
The matrices defining the wrench transformation are
\begin{equation}
\label{eq:ad_q_bci_Bc}
    \mathrm{Ad}^{\top}_{q^{-1}_{b,c_i}}
    =
    \begin{bmatrix}
        R_{b,c_i} & 0\\
        -R_{b,c_i}\hat{p}_{c_i,b} & R_{b,c_i}
    \end{bmatrix},
    \qquad
    B_c=
    \begin{bmatrix}
        I_{3\times3}\\
        0_{3\times3}
    \end{bmatrix},
\end{equation}
where $\hat{p}_{c_i,b}\in\mathrm{so}(3)$ is the skew-symmetric matrix associated with the contact-point position.

Each contact force is constrained to lie within its friction cone. For optimization, the circular cone is conservatively approximated by an inscribed polyhedral cone with $k$ unit generators $\hat f_{c_i,j}$. For $k=4$, the approximation is pyramidal, as shown
in Fig.~\ref{fig:system-description}. Thus,
\begin{equation}
\label{eq:linearizedFC}
    FC_{c_i}
    =
    \left\{
    f_{c_i}\in\mathbb{R}^3
    \,\middle|\,
    f_{c_i}=\sum_{j=1}^{k}
    \lambda_{c_i,j}\hat f_{c_i,j},
    \quad
    \lambda_{c_i,j}\geq0
    \right\}.
\end{equation}
Stacking the contact forces gives
\begin{equation}
\label{eq:fc}
    F_c=\hat F_c\Lambda,
    \qquad
    \Lambda=
    \begin{bmatrix}
        \lambda_{1,1}&\cdots&\lambda_{n_c,k}
    \end{bmatrix}^{\top}\geq0,
\end{equation}
where
\begin{equation}
    \hat F_c
    =
    \operatorname{diag}
    \left(\hat F_{c,1},\ldots,\hat F_{c,n_c}\right),
    \qquad
    \hat F_{c,i}
    =
    \begin{bmatrix}
        \hat f_{c_i,1}&\cdots&\hat f_{c_i,k}
    \end{bmatrix}.
\end{equation}
Consequently, the admissible body wrench is expressed as
\begin{equation}\label{lambda_positive}
    \mathcal{F}_b=G\hat F_c\Lambda,
    \qquad
    \Lambda\geq0.
\end{equation}
This model allows to keep the object non-sliding conditions by enforcing $\Lambda\geq0$.

\section{Methodology} \label{sec:method}

\subsection{Robust Trajectory Generation}\label{sec:Robust Trajectory Generation}
\begin{figure}[t!]
    \centering
    \resizebox{0.85\columnwidth}{!}{%
        \input{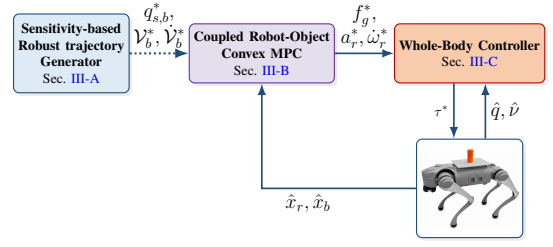}
    }
    \caption{Conceptual block scheme of the proposed control architecture. The
sensitivity-based trajectory generator provides the desired object
pose, twist, and its derivative to the coupled
robot-object convex MPC. Using these references and the estimated robot and object states, the MPC computes the desired ground reaction forces and base accelerations. The whole-body controller maps these
commands and the measured generalized state into the joint torque command applied to the quadruped. 
} 
    \label{fig:method_arch}
\end{figure}

The proposed framework is summarized in Fig.~\ref{fig:method_arch}. The trajectory generator parametrizes the translational and rotational components of the desired object trajectory by eighth-degree polynomials. Their coefficients are stacked in
$\alpha\in\mathbb{R}^{n_\alpha}$, which constitutes the optimization
variable, while the corresponding velocities and accelerations are obtained from the first and second polynomial derivatives.

Following~\cite{giordano2018trajectory}, we seek the coefficients $\alpha^\star$ that minimize the sensitivity of the dynamical system in~\eqref{eq:model} to parametric variations. Compared with the initial study in~\cite{teimoorzadeh2025robust}, the present work considers a broader set of uncertainties and incorporates the physical and kinematic constraints of the quadruped into the optimization.

Let $\mathcal{T}=[t_0,t_f]$, $\mathcal{I}=\{x,y,z\}$, and $\mathcal{J}=\{\phi,\theta,\psi\}$. The
trajectory optimization is formulated as
\begin{equation}
\label{eq:sensitivity_eq}
\begin{aligned}
\alpha^\star=\arg\min_{\alpha}\quad
&\int_{t_0}^{t_f}\|\Pi(t)\|_F^2\,\mathrm{d}t\\
\mathrm{s.t.}\quad
&y^*(\alpha,t_0)=y_0,\qquad
 y^*(\alpha,t_f)=y_f,\\
&\Lambda(t)-r_{\Lambda}(t)\succeq0,\\
&\dot{x}(t)=f\bigl(x(t),\Lambda(t),p\bigr),\\
&z_{\min}\leq z_b(t)\leq z_{\max},\\
|v_{b,i}(t)|&\leq v_{b,i,\max},\qquad
|\omega_{b,j}(t)|\leq\omega_{b,j,\max},\\
|\dot v_{b,i}(t)|&\leq\dot v_{b,i,\max},\qquad
 |\dot\omega_{b,j}(t)|\leq\dot\omega_{b,j,\max}, \\
&i\in\mathcal I,\qquad j\in\mathcal J,\qquad t\in\mathcal T.
\end{aligned}
\end{equation}

The objective quantifies the sensitivity magnitude using the squared Frobenius norm all over the trajectory.  
The uncertain parameters are collected in $p\in\mathbb{R}^{n_p}$ which includes the inertial parameters as well as the friction coefficient between the object and the robot's back-mounted platform.

All path constraints are enforced for every $t\in\mathcal{T}$,
and $y_0 = \left(q_{s,b_0},\mathcal{V}_{b_0},
\dot{\mathcal{V}}_{b_0}\right)$ and
$y_f = \left(q_{s,b_f},\mathcal{V}_{b_f},
\dot{\mathcal{V}}_{b_f}\right)$ denote the prescribed initial and final conditions, respectively. 
Moreover, $r_{\Lambda_i}(t)$ denotes the tube radius associated with
the $i$-th contact coefficient $\lambda_i(t)$, computed according
to~\eqref{eq:radius}.
Thus, the constraint
$\Lambda(t)-r_\Lambda(t)\succeq0$ ensures that the lower boundary of the predicted input tube remains inside the polyhedral friction cone.
The numerical values of the kinematic bounds imposed during the robust trajectory optimization are reported in Sec.~\ref{sec:simulation}. 

The closed-loop dynamics $\dot{x}(t)$ appearing in~\eqref{eq:sensitivity_eq} is obtained by combining the object dynamics in~\eqref{eq:object_dynamics} with the trajectory tracking control law introduced below. To this end, the object dynamics can be expressed by
\begin{equation}\label{eq:f}
\begin{aligned}
    	\dot x=
	\begin{cases}
		\dot{q}_{s,b} = \dot{q}_{s,b}\\
		\dot{\calV}_b = M_b^{-1}(q) \left(G\hat F_c \Lambda - C_b({\calV}_b){\calV}_b - N_b\left(R_{s,b}\right)\right)
	\end{cases}.
\end{aligned}
\end{equation}
Taking inspiration by previous works on the topic (e.g.~\cite{selvaggio2023non}), a suitable control law to track a desired trajectory $y^*(t) = \left(q^*_{s,b}, \calV^*_{b}, \dot{\calV}^*_{b}\right)$ can be devised as follows 
\begin{equation}\label{eq:control}
    \begin{aligned}
          \Lambda = \left(G\hat{F}_c\right)^\dagger\Big(&C_b(\mathcal{V}_b)\mathcal{V}_b + N_b\left(R_{s,b}\right) + M_b\left(\dot{\mathcal{V}}^{*}_b + \right. \\ + K_d 
          & \left. \left(\mathcal{V}_b^{*}(t) - \mathcal{V}_b\right) + K_p\left(q^*_{s,b} - q_{s,b}\right)\right)\Big).
    \end{aligned}
\end{equation}
Equation~\eqref{eq:f} uses the actual parameters $p$, while the controller in~\eqref{eq:control} uses their nominal values $p_n$.
Thus,
Collectively, these equations define the object dynamics as
\begin{equation}
\dot x=f(x,\Lambda,p).
\label{eq}
\end{equation}
which defines the closed-loop dynamics.
The non-sliding constraint uses the sensitivity tubes
in~\eqref{eq:tube} to keep the contact forces within the friction cones
under the considered parameter variations. Thus, the optimized
trajectory reduces sensitivity to object uncertainty while maintaining
contact feasibility.
\subsection{Coupled Robot-Object Convex MPC Modeling}~\label{sec.convex}
The proposed control architecture combines a coupled robot--object
centroidal MPC with a whole-body inverse-dynamics QP. The MPC jointly
optimizes the ground-reaction forces and the object--platform contact
wrench, while the whole-body QP converts the resulting centroidal
commands into joint torques.
Within the MPC, the quadruped is represented using the single-rigid-body dynamics formulation 
\begin{equation}
\label{eq:robot_srbd_dynamics}
\begin{aligned}
    m_r\dot v_r
    &=
    \sum_{i=1}^{n_f}f_i
    -f_{s,b}
    +m_rg,\\
    \frac{\mathrm{d}}{\mathrm{d}t}
    \left(I_r\omega_r\right)
    &=
    \sum_{i=1}^{n_f}r_i\times f_i
    -\tau_{s,b}
    +(p_b-p_r)\times(-f_{s,b}),\\
    \dot R_r
    &=
    [\omega_r]_\times R_r.
\end{aligned}
\end{equation}
where $m_r$ and $I_r$ denote the robot mass and its centroidal inertia
expressed in the world frame, respectively.

Using the contact-wrench mapping introduced in the previous section,
the object wrench expressed in the spatial frame is
\begin{equation}
\label{eq:object_wrench_spatial}
    \mathcal F_{s,b}
    =
    \begin{bmatrix}
        R_{s,b} & 0\\
        0 & R_{s,b}
    \end{bmatrix}
    G\hat F_c\Lambda
    =
    H_{s,b}\Lambda,
\end{equation}
where $ H_{s,b}$ maps the nonnegative contact coefficients $\Lambda$
to the object wrench. Its equal-and-opposite reaction is included in
the robot dynamics.

The robot prediction state, stacked ground-reaction forces and the object prediction state are defined as
\begin{equation}
\begin{aligned}
    x_r &=
    \begin{bmatrix}
        p_r^\top &
        \Theta_r^\top &
        v_r^\top &
        \omega_r^\top
    \end{bmatrix}^{\top}
    \in\mathbb{R}^{12},\\
    f_g &=
    \begin{bmatrix}
        f_1^\top &
        \cdots &
        f_{n_f}^\top
    \end{bmatrix}^{\top}
    \in\mathbb{R}^{3n_f},\\
    x_b &=
    \begin{bmatrix}
        p_b^\top &
        \Theta_b^\top &
        v_b^\top &
        \omega_b^\top
    \end{bmatrix}^{\top}
    \in\mathbb{R}^{12}.\\
\end{aligned}
\end{equation}
Linearizing the configuration dependent terms of the model in~\eqref{eq:robot_srbd_dynamics} along the nominal trajectory and discretizing them with the MPC sampling time gives
\begin{equation}
\label{eq:robot_discrete_ltv}
    x_{r,k+1}
    =
    A_{r,k}x_{r,k}
    +B_{g,r,k}f_{g,k}
    +B_{\Lambda,r,k}\Lambda_k
    +d_{r,k}.
\end{equation}
\begin{equation}
    x_k=
    \begin{bmatrix}
        x_{r,k}\\
        x_{b,k}
    \end{bmatrix}
    \in\mathbb{R}^{24},
    \qquad
    u_k=
    \begin{bmatrix}
        f_{g,k}\\
        \Lambda_k
    \end{bmatrix},
\end{equation}
\begin{equation}
\label{eq:coupled_prediction_model}
    x_{k+1}=A_kx_k+B_ku_k+d_k.
\end{equation}
The object and platform motions are coupled through the softened
acceleration-level conditions
\begin{align}
    a_{b,k}
    &=
    a_{r,k}
    +\dot\omega_{r,k}\times\bar\rho_k
    +\bar\omega_{r,k}\times
    \left(\bar\omega_{r,k}\times\bar\rho_k\right)
    +s_{a,k},
    \label{eq:translational_coupling}\\
    \dot\omega_{b,k}
    &=
    \dot\omega_{r,k}+s_{\omega,k},
    \label{eq:rotational_coupling}
\end{align}
where $\bar\rho_k=\bar p_{b,k}-\bar p_{r,k}$ is evaluated along the
nominal trajectory and
$s_k=[s_{a,k}^\top,s_{\omega,k}^\top]^\top\in\mathbb{R}^{6}$.
The slack variable allows deviations from perfect object--platform motion coupling.

At each control update, the following finite-horizon QP is solved:
\begin{equation}
\label{eq:coupled_mpc}
\begin{aligned}
\min_{\{x_k,u_k,s_k\}}\quad
&\sum_{k=0}^{N-1}
\left(
\|x_k-x_k^d\|_Q^2
+\|u_k\|_R^2
+w_s\|s_k\|_2^2
\right)\\
\mathrm{s.t.}\quad
&x_0=\hat x,\\
&x_{k+1}=A_kx_k+B_ku_k+d_k,\\
&\Lambda_k\succeq0,\qquad
f_{g,k}\in\mathcal F_{g,k},\\
&\eqref{eq:translational_coupling},
\quad\eqref{eq:rotational_coupling},\\
&k=0,\ldots,N-1.
\end{aligned}
\end{equation}
where $\mathcal{F}_{g,k}$ denotes the feasible ground reaction force set, including friction constraints and zero forces at swing feet.

\subsection{Whole-Body Inverse-Dynamics Control}~\label{sec.wbc}
The MPC centroidal commands are tracked by a whole-body inverse-dynamics QP that considers the full floating-base dynamics and enforces contact and actuator constraints. Let $q$ denote the robot configuration and $\nu\in\mathbb{R}^{18}$ its generalized velocity. The QP decision variable is
\begin{equation}
\label{eq:wbc_variables}
    \chi=
    \begin{bmatrix}
        \dot\nu^\top&
        \tau^\top&
        f_g^\top
    \end{bmatrix}^{\top},
\end{equation}
where $\dot\nu\in\mathbb{R}^{18}$ is the generalized acceleration, $\tau\in\mathbb{R}^{12}$ contains the actuated joint torques, and $f_g\in\mathbb{R}^{3n_f}$ is the stacked vector of ground-reaction forces.

The full floating-base dynamics are imposed as an equality constraint:
\begin{equation}
\label{eq:wbc_dynamics}
    M(q)\dot\nu+h(q,\nu)
    =
    S^\top\tau
    +J_g^\top(q)f_g
    +\tau_{\mathrm{ext},b},
\end{equation}
where $M(q)$ is the mass matrix, $h(q,\nu)$ contains the Coriolis, centrifugal, and gravitational terms, and $S$ selects the actuated joints. Moreover, $J_g(q)$ is the stacked foot-contact Jacobian, and $\tau_{\mathrm{ext},b}\in\mathbb{R}^{18}$ is the generalized force exerted by the payload on the robot.

For the stance feet, zero contact point acceleration is enforced through
\begin{equation}
\label{eq:stance_acceleration_constraint}
    J_{\mathrm{st}}(q)\dot\nu
    +\dot J_{\mathrm{st}}(q,\nu)\nu
    =0,
\end{equation}
where $J_{\mathrm{st}}$ is the contact Jacobians of the active stance feet.
The ground-reaction forces are additionally constrained by
\begin{equation}
\label{eq:wbc_contact_constraints}
    f_i\in FC_{g,i},
    \quad i\in\mathcal I_{\mathrm{st}},
    \qquad
    f_i=0,
    \quad i\in\mathcal I_{\mathrm{sw}},
\end{equation}
where $\mathcal I_{\mathrm{st}}$ and $\mathcal I_{\mathrm{sw}}$ denote
the sets of stance and swing feet, respectively, and $FC_{g,i}$ is the
linearized ground-friction cone of the $i$-th foot. The actuator limits
are imposed as
\begin{equation}
\label{eq:torque_limits}
    \tau_{\min}\leq\tau\leq\tau_{\max},
\end{equation}
where the inequalities are applied component-wise.
The MPC provides the desired robot acceleration
\begin{equation}
    a_r^d=
    \begin{bmatrix}
        (a_{\mathrm{lin}}^d)^\top&
        (\dot\omega_r^d)^\top
    \end{bmatrix}^{\top}
    \in\mathbb{R}^{6},
\end{equation}
together with the desired ground-reaction forces $f_g^d$. The
corresponding whole-body task errors are
\begin{align}
    e_r
    &=
    J_r(q)\dot\nu
    +\dot J_r(q,\nu)\nu-a_r^d,
    \label{eq:base_acceleration_error}\\
    e_{\mathrm{sw}}
    &=
    J_{\mathrm{sw}}(q)\dot\nu
    +\dot J_{\mathrm{sw}}(q,\nu)\nu
    -a_{\mathrm{sw}}^d,
    \label{eq:swing_acceleration_error}\\
    e_f
    &=f_g-f_g^d.
    \label{eq:force_tracking_error}
\end{align}
where $J_r$ combines the robot CoM linear-velocity Jacobian and the base angular velocity Jacobian, while $J_{\mathrm{sw}}$ stacks the swing foot Jacobians.

At each control step, the whole-body command is obtained by solving
\begin{equation}
\label{eq:wbc_qp}
\begin{aligned}
    \min_{\chi}\quad
    &
    \|e_r\|_{W_r}^{2}
    +\|e_{\mathrm{sw}}\|_{W_{\mathrm{sw}}}^{2}
    +\|e_f\|_{W_f}^{2}
    +\|\tau\|_{W_\tau}^{2}
    +\|\dot\nu\|_{W_{\dot\nu}}^{2}
    \\
    \mathrm{s.t.}\quad
    &
    \eqref{eq:wbc_dynamics},
    \quad
    \eqref{eq:stance_acceleration_constraint},
    \quad
    \eqref{eq:wbc_contact_constraints},
    \quad
    \eqref{eq:torque_limits}.
\end{aligned}
\end{equation}
Here, $\|e\|_W^2=e^\top We$, and the positive semidefinite weighting matrices determine the relative importance of the tracking and regularization objectives. The optimal torque $\tau^\star$ is applied
to the robot at each control update.

\section{Simulation and Hardware Validation} \label{sec:simulation}
In this section, we present the simulation and hardware experiments
conducted to validate the proposed framework for robust object
transportation using a quadruped robot\footnote{The videos of the experiments can be found on the project website: \\ \texttt{https://quadtransport.github.io/icra/}}.
A cylindrical object with a nominal mass of $0.07$~kg, radius of
$10$~cm, and height of $20$~cm is placed on a flat platform firmly
mounted on the quadruped's back. The object remains free to slide and
is transported solely through frictional interaction with the
platform. The nominal friction coefficient in simulation environment is set to $\mu=0.52$.

The optimization problem in~\eqref{eq:sensitivity_eq} is implemented
in MATLAB using CasADi~\cite{andersson2019casadi}. Three trajectories
with different travel distances are generated from the initial point
$p_{s,b_0}=(0,0,0.38)^\top$~m
to the final point
$
p_{s,b_f}\in
\left\{
(2,0,0.38)^\top,\,
(2.2,0,0.38)^\top,\,
(2.4,0,0.38)^\top
\right\}\text{m}.
$
All trajectories have a fixed duration of $5$~s, making the longer
distances increasingly dynamic. The trajectory generation parameters
are summarized in Table~\ref{tab:trajectory_constraints}. The
optimization takes an average of $65$~s on a desktop computer equipped
with an Intel Core i9-10900K processor.
\begin{table}[!hbt]
    \centering
\caption{Trajectory generation constraints and control parameters}
    \label{tab:trajectory_constraints}
    \begin{tabular}{lcc}
\hline
Object CoM height
& $z_b$
& $0.35 \leq z_b \leq 0.40~\mathrm{m}$ \\

Forward velocity
& $v_{b,x}$
& $|v_{b,x}| \leq 1.0~[\mathrm{m/s}]$ \\

Lateral velocity
& $v_{b,y}$
& $|v_{b,y}| \leq 0.05~[\mathrm{m/s}]$ \\

Vertical velocity
& $v_{b,z}$
& $|v_{b,z}| \leq 0.2~[\mathrm{m/s}]$ \\

   Control Parameters 
   & $K_p$ = diag$(250, 15)$ 
   & $K_d$ = diag$(15,0.2)$ \\
        \hline
    \end{tabular}
\end{table}

The desired object trajectory is parameterized by eighth-degree
polynomials, whose coefficients are collected in $\alpha$. These
polynomials determine the corresponding pose, twist, and acceleration trajectories, with the orientation quaternion obtained by integrating the angular velocity. The prescribed boundary conditions, including zero twist and acceleration at both endpoints, are expressed as $ M\alpha=d$,
where $M$ evaluates the polynomial basis and its derivatives at $t_0$ and $t_f$, and $d$ contains the prescribed boundary values. The coefficient vector
$
    \alpha_0=M^\dagger d_0
$
is used as the initial guess for the CasADi optimization solved with IPOPT as the minimum norm polynomial satisfying the prescribed position, orientation, velocity and acceleration in the boundary conditions.

\begin{figure}[t]
    \centering
    \includegraphics[width=\linewidth]{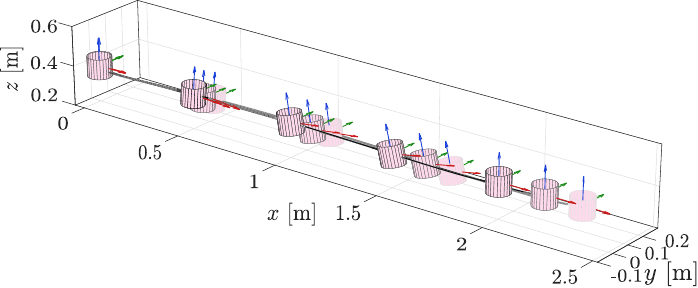}
    \caption{3D visualization of the optimized robust object trajectories over a $5$~s horizon for travel distances of $2$, $2.2$, and $2.4$~m, shown in black, gray, and light gray, respectively. The object orientations are illustrated at intervals of
$0.25$ along the travel distance.}
    \label{fig:3D_trajectories}
\end{figure}

The optimized 3D trajectories for the three target distances are shown in Fig.~\ref{fig:3D_trajectories}. Despite the increasingly dynamic motion required over longer distances, the optimizer finds a feasible solution in all cases. 
As observed, the optimized trajectories deviate slightly from a straight line, exhibiting a bell-shaped vertical profile combined with backward and forward pitch rotations. From a physical perspective, the bell-shaped vertical motion can be interpreted as increasing the normal component of the contact forces during upward acceleration. However, due to the robot’s limited admissible workspace in the vertical direction, as imposed by the constraints in~\eqref{eq:sensitivity_eq}, the observed pitch motion may provide a means of compensating for this limitation. 
\begin{figure}[t]
    \centering
    \includegraphics[width=0.45\textwidth]{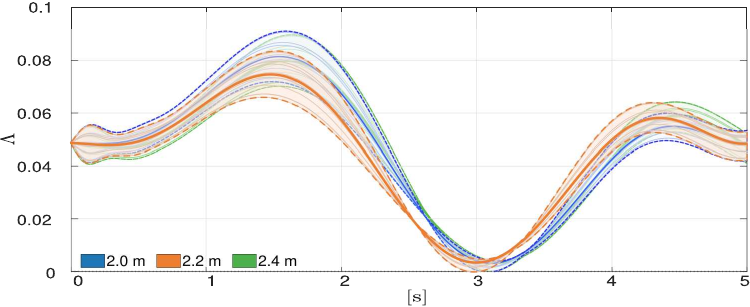}
    \caption{Time evolution of a representative $\Lambda$ for the optimized trajectories with travel distances of $2$, $2.2$, and $2.4$~m. The thick solid curves represent the nominal solutions, the thin curves show the Monte Carlo realizations under $\pm10\%$ perturbations of the object inertial parameters, and the dashed curves indicate the corresponding envelopes.}
    \label{fig:Lambda_tubes}
\end{figure}

To evaluate the robustness of the generated trajectories and the accuracy of the computed sensitivity tubes, a Monte Carlo analysis with  $30$ realizations was performed by independently varying the object uncertain parameters within $\pm10\%$ of their nominal values. The analysis was repeated for each of the three optimized trajectories. Figure~\ref{fig:Lambda_tubes} shows the resulting evolution of the contact coefficients $\Lambda$ together with their sensitivity-based envelopes that remain within the predicted tubes satisfying~\eqref{lambda_positive}.

The optimized trajectories are then evaluated using \(30\) Monte Carlo realizations under perturbations of the nominal object parameters. The evaluations are performed in a MuJoCo environment calibrated using motion-capture data, with the quadruped dynamics and controller included in the closed loop as in Fig.~\ref{fig:method_arch}. Performance is assessed using two metrics: the RMS error between the actual and
desired robot CoM poses, $p_{s,r}$ and $p_{s,r}^{*}$, respectively, and the final
planar object displacement
$s_{xy}=\|p_{\mathcal P,b_f}^{xy}-p_{\mathcal P,b_0}^{xy}\|_2$.
The superscript $*$ denotes the desired pose, while the subscripts $0$ and $f$
denote the initial and final object instants, respectively. 

\begin{figure}[!tbp]
    \centering
    \includegraphics[width=0.47\textwidth]{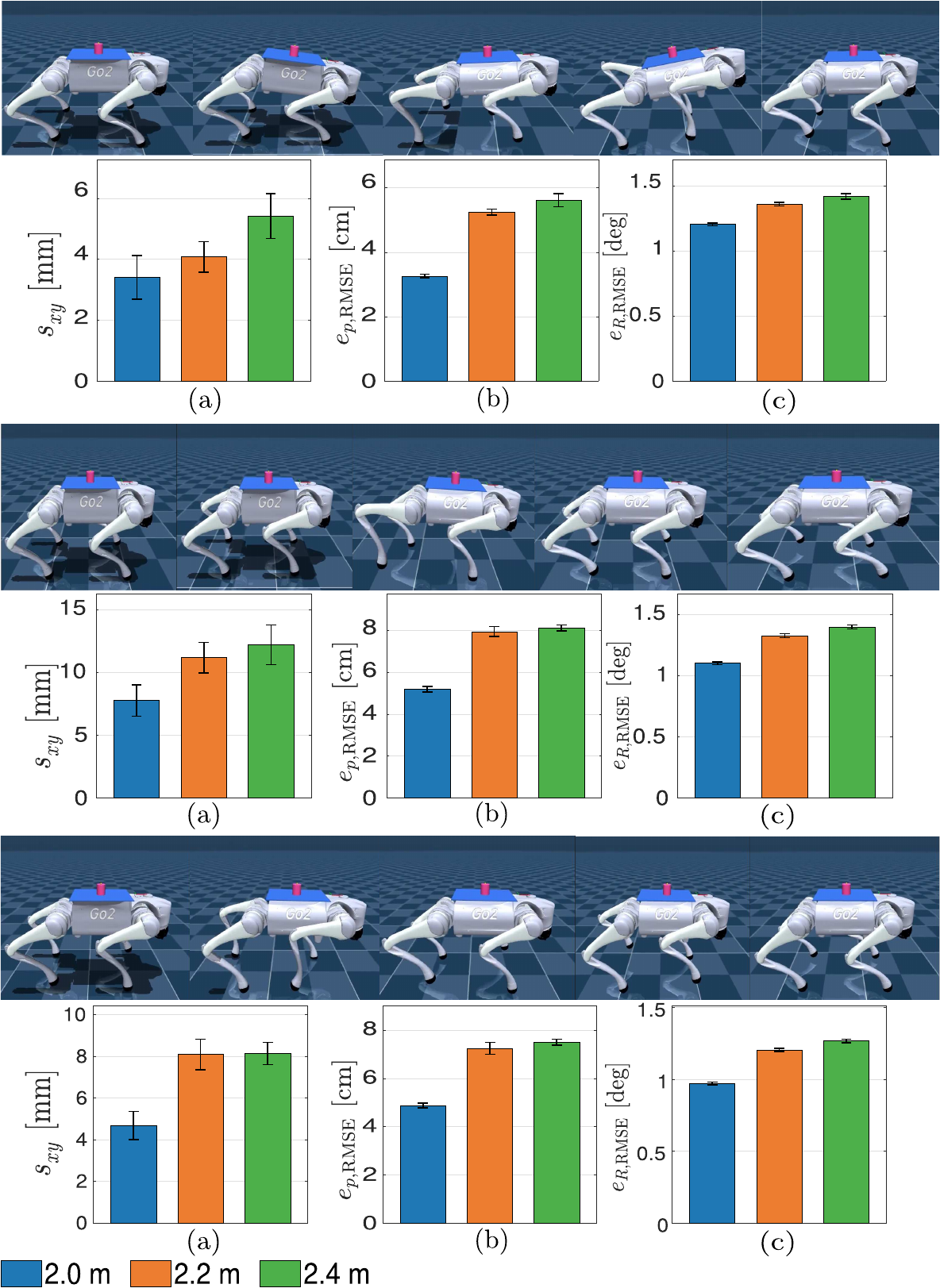}
    \caption{Monte Carlo comparison of the optimized robust (top), fixed-orientation
(middle), and straight-line (bottom) trajectories performed in the MuJoCo simulation environment. Each row shows
representative snapshots and (a) robot CoM tracking error, (b) object
RMS translation, and (c) object RMS orientation deviation. Blue,
orange, and green denote $2$, $2.2$, and $2.4$~m, respectively; error
bars show the variation across realizations.}
    \label{fig:robust_trajectories}
\end{figure}

To explore the roles of vertical motion and pitch rotation in optimized trajectories, Fig.~\ref{fig:robust_trajectories} compares the optimized trajectories with fixed-orientation and straight-line baselines through Monte Carlo simulations. Comparing the same colored bars across the three rows
shows that the optimized trajectory achieves the lowest robot CoM
tracking error and object translational deviation for all travel
distances upon changes in inertial parameters. Although the errors increase with distance due to the more dynamic motion, the improvement over both baselines is maintained for the $2.0$, $2.2$, and $2.4$~m cases. The object orientation deviation remains relatively high because of the increased motion dynamics of the optimized trajectory.

\begin{figure}[!tbp]
    \centering
    \includegraphics[width=0.4\textwidth]{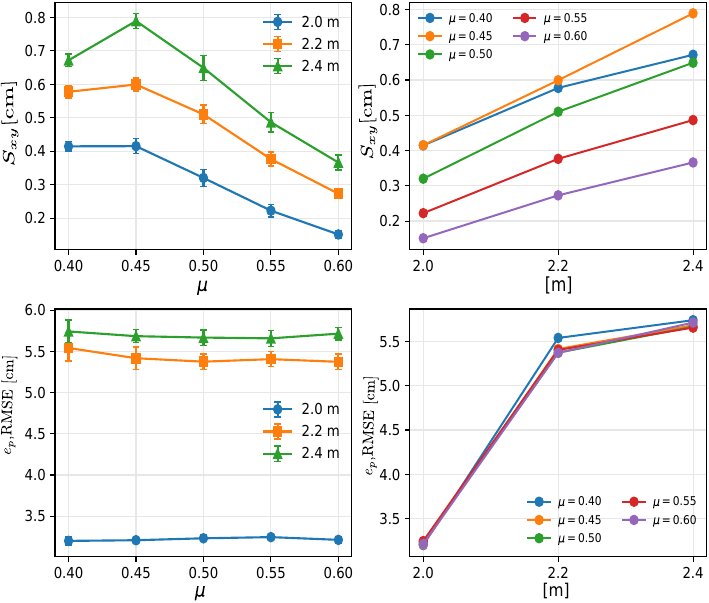}
    \caption{Monte Carlo evaluation of the optimized trajectories under friction variations, with $30$ realizations per coefficient.}
    \label{fig:fric_mass_}
\end{figure}

Figure~\ref{fig:fric_mass_} represents the Monte Carlo evaluation of the optimal trajectories under variations in the object-platform friction coefficient and object mass. Five friction coefficients and
three travel distances were tested using $30$ mass realizations for
each combination, resulting in $450$ simulations. Increasing friction generally reduces object sliding, with a greater effect for longer trajectories. The robot CoM tracking error increases with travel distance due to the more demanding motion but remains nearly unaffected by friction variations.

\begin{figure}[!h] 
    \centering
    \includegraphics[width=0.47\textwidth]{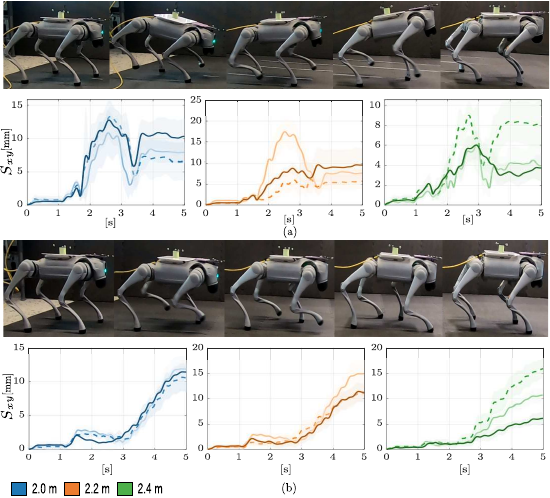}
    \caption{ Time evolution of the object sliding for (a) the optimized trajectory and (b) straight-line baseline. The dashed curves represent the nominal case, while the dark and light curves correspond to \(+10\%\) and \(-10\%\) variations in the nominal object mass, respectively.}
    \label{fig:time_error}
\end{figure}

To validate the simulation results, the experiments were repeated on
the hardware setup, as shown in Fig.~\ref{fig:time_error}. The
optimized and straight line baseline trajectories used the same travel
distances and durations as in the simulation. Each trajectory was executed
ten times to reduce the effect of real-world variations. The
experiments considered the nominal object mass $m_n$ and the two
extreme cases $m_n\pm10\%$. With three travel distances, two trajectory
types, three mass values, and ten repetitions, a total of $180$ trials
were performed.

\hfill

Figure~\ref{fig:time_error}(a) reports the object sliding displacement obtained
with the optimized trajectory, while Fig.~\ref{fig:time_error}(b)
shows the corresponding straight line results. The shaded
regions indicate the standard deviation across the ten repetitions.
The dashed curve represents the nominal mass, whereas the dark and
light solid curves correspond to the two worst-case mass variations,
$+10\%$ and $-10\%$, respectively.
Due to limitations of the hardware platform, the robot's default controller was used during the experiments instead of the
proposed whole-body controller. Since the same controller was employed
for all trials, the comparison isolates the effect of the trajectory
design. Due to this limitation, the optimized trajectories exhibit increased sliding around the midpoint but recover toward the end of the motion, resulting in lower final object displacement. 

\section{Conclusions} \label{sec:conclusions}
In this paper, we proposed a framework for robust nonprehensile object transportation using a platform mounted on the back of a quadruped. The proposed closed-loop sensitivity-based trajectory optimization approach generates trajectories with reduced sensitivity to uncertainties in the object's inertial parameters and the friction coefficient. The optimized trajectories combine a bell-shaped vertical motion with backward and forward pitch rotations, effectively exploiting the quadruped’s limited workspace. Extensive MuJoCo evaluations show lower robot CoM tracking errors and reduced object sliding compared with fixed-orientation and straight-line baselines. Hardware experiments using the robot's native controller further demonstrate the benefits of the optimized trajectories under payload mass variations. Future work will implement the complete control framework on hardware and extend it to broader object geometries and larger parameter uncertainties.

\bibliographystyle{IEEEtran}
\bibliography{ref}

\end{document}